\documentclass[10pt]{article}
\usepackage{multicol}
\usepackage{times}
\usepackage[T3,OT2,T1]{fontenc} 
\usepackage[tone]{tipa}
\usepackage{tipx}
\usepackage{bm}
\usepackage{color}
\usepackage[colorlinks,linkcolor=blue]{hyperref}
\usepackage{tikz-cd}
\usepackage{amssymb}
\usepackage{tcolorbox}
\usepackage{amsmath}
\usepackage{enumerate}
\usepackage{algorithm}
\usepackage[noend]{algpseudocode}
\usepackage{amsthm}
\usepackage{mathrsfs}
\usepackage{geometry}
\usepackage[colorlinks, linkcolor=blue]{hyperref}
\usepackage{verbatim}
\usepackage{enumitem}
\usepackage{bbm}
\usepackage{theoremref}
\usepackage{booktabs}
\usepackage{listings}
\usepackage{graphicx}
\usepackage{subfigure}
\usepackage{caption}
\usepackage[frozencache=true,cachedir=_minted-article]{minted}

\setminted[python]{
    breaklines, framesep=2mm, fontsize=\footnotesize, numbersep=5pt
}

\newcommand{\N}{\mathbb{N}}
\newcommand{\E}{\mathbb{E}}

\newcommand{\R}{\mathbb{R}}

\renewcommand{\P}{\mathbb{P}}

\newcommand{\sgn}{\text{sgn}}

\newcommand{\dd}{\mathrm{d}}

\DeclareMathOperator{\tr}{tr}

\newcommand{\T}{ ^\text{T} }

\newtheorem{lemma}{Lemma}

\newtheorem{thm}{Theorem}

\title{Cramer Type Distances for Learning Gaussian Mixture Models by Gradient Descent}
\author{ \textbf{Ruichong Zhang}\\
Tsinghua University\\
\texttt{zhangrc20@mails.tsinghua.edu.cn} \\ }

\begin{document}
\maketitle
\begin{abstract}

    The learning of Gaussian Mixture Models (also referred to simply as GMMs) plays an important role in machine learning. Known for their expressiveness and interpretability, Gaussian mixture models have a wide range of applications, from statistics, computer vision to distributional reinforcement learning.
    However, as of today, few known algorithms can fit or learn these models, some of which include Expectation-Maximization algorithms and Sliced Wasserstein Distance. Even fewer algorithms are compatible with gradient descent, the common learning process for neural networks.
    
    In this paper, we derive a closed formula of two GMMs in the univariate, one-dimensional case, then propose a distance function called Sliced Cram\'er 2-distance for learning general multivariate GMMs. Our approach has several advantages over many previous methods. First, it has a closed-form expression for the univariate case and is easy to compute and implement using common machine learning libraries (e.g., PyTorch and TensorFlow). Second, it is compatible with gradient descent, which enables us to integrate GMMs with neural networks seamlessly. Third, it can fit a GMM not only to a set of data points, but also to another GMM directly, without sampling from the target model. And fourth, it has some theoretical guarantees like global gradient boundedness and unbiased sampling gradient. These features are especially useful for distributional reinforcement learning and Deep Q Networks, where the goal is to learn a distribution over future rewards. We will also construct a Gaussian Mixture Distributional Deep Q Network as a toy example to demonstrate its effectiveness. Compared with previous models, this model is parameter efficient in terms of representing a distribution and possesses better interpretability.
\end{abstract}

\section{Introduction}\ 

Gaussian Mixture Models, also known as Mixture of Gaussians, sometimes abbreviated as GMMs or MoGs, are renowned for their expressiveness and interpretability, and apply in fields like signal processing \cite{plataniotis2000gaussian}, generative adversarial nets \cite{ben-yosef2018gaussian}, distributional reinforcement learning \cite{choi2019distributional}, Autoencoders for image generation \cite{smieja2020segma} and much more. 
The learning or fitting of GMMs, or estimating the parameters of GMM given the data distribution, has long been a major concern in the field of machine learning. 
The most famous approaches include the Expectation-Maximization (EM) algorithm, which is equivalent to minimizing the Negative Log Likelihood loss, but might suffer heavily from local optima problem \cite{dempster1977maximum,tipping1999mixtures}, or might seem powerless dealing with neural networks; and gradient descent based methods, like the sliced Wasserstein distance \cite{kolouri2017sliced} or Wasserstein-Fischer-Rao gradient flow \cite{yan2023learning}, which generally performs better than expectation or likelihood based iteration algorithms, while being compatible with neural network learning. 

The Cram\'er 2-distance \cite{cramer1936some}, or known as the $L^2$ distance between cumulative distribution functions of two univariate random variables, is used to fit probability distributions, also applicable to distributional reinforcement learning \cite{lheritier2022cramer}. As an alternative to the Wasserstein distance, it is known to enjoy certain key properties, like unbiased sampling gradient and contraction in the distributional Bellman operator \cite{bellemare2017cramer}, 

The Sliced Cram\'er 2-distance, also known as the Cram\'er-Wold distance \cite{kolouri2020sliced,knop2019cramer}, is considered as a natural generalization of Cram\'er 2-distance for random vectors or distributions in higher dimensional spaces. Guaranteed by the Cram\'er-Wold theorem, it is calculated by taking projections of distributions along all unit vectors on the sphere, and integrating up the 1D Cram\'er distance of the projected distributions. The closed form formula of Sliced Cram\'er 2-distance between spherical (isotropic) Gaussians have been proposed, using hypergeometric functions. \cite{smieja2020segma}

Although these Cram\'er type distances have been applied to GMM learning, the main purpose of our work is a bit different. Our work mainly focuses on the following points:
\begin{itemize}
    \item Derive a closed formula for the Cram\'er 2-distance for univariate (1D) GMM learning, which is accessible directly through common machine learning libraries.
    \item Use the Sliced Cram\'er 2-distance for general multivariate GMM learning, applicable to general mixtures of anisotropic Gaussians.
    \item Offer detailed formula derivation processes and proofs, including avoidance of gradient explosion and unbiased sampling gradients.
    \item Conduct some basic experiments to demonstrate the feasibility of our approaches.
\end{itemize}

\section{Preliminaries About GMMs}\ 

In this section, we will go over some definitions that is crucial to our formulation of the theory, as well as the related previous works.
\subsection{Multivariate Gaussian Distribution}\ 

The Gaussian distribution is of central importance in the theory of probability and statistics. It is known from the central limit theorem that in most situations, standard sampled mean of independent, identically distributed random variables tends to a Gaussian distribution. 

Let $m \in \N^+$ be a positive integer. In all cases below, we denote by $m$ the dimension number.

A multivariate Gaussian distribution (also called Gaussian random vector, or $m$-dimensional Gaussian distribution) in $\R^m$ is defined as 
$\mathcal{N}(\bm{\mu}, \bm{\Sigma})$ where $\bm{\mu} \in \R^{m}$ is a vector, and $\bm{\Sigma} \in M_m(\R)$ is a positive-definite matrix. 
The probability density function (PDF) is
$$\mathcal{N}(\bm{\mu}, \bm{\Sigma})(\bm{x}) = \frac{1}{(2\pi)^{m/2}\sqrt{\det(\bm{\Sigma})}} \exp\left(-\frac{(\bm{x}-\bm{\mu})\T \bm{\Sigma}^{-1} (\bm{x}-\bm{\mu})}{2}\right) $$
This Gaussian distribution is called spherical, or isotropic, if $\bm{\Sigma}$ is a multiple of the identity matrix $\bm{I}_m$, and anisotropic if otherwise.

When $m=1$, we obtain the univariate case:
$$\mathcal{N}(\mu, \sigma^2)({x}) = \frac{1}{\sqrt{2\pi}\sigma} \exp\left(-\frac{({x}-{\mu})^2}{2\sigma^2}\right) $$
Which has expectation $\mu$ and standard deviation $\sigma$. When $\sigma = 0$, the distribution is degenerate as a single-point distribution. All univariate Gaussians are isotropic.

A property of the multivariate Gaussian distribution is that its inner product with another vector is a univariate Gaussian random variable \cite{davar2014gaussian}.

For a general multivariate Gaussian distribution, the $\bm{\Sigma} \in M_m(\R)$ is not guaranteed to be strictly positive definite, (i.e., $\text{rank}(\bm{\Sigma}) < m $). Thus, the probability distribution function may fail to exist in the common sense. 
However, the projection of $\mathcal{N}(\bm{\mu}, \bm{\Sigma})$ along a certain unit vector $\bm{a}$ exists and is still a Gaussian distribution. 
If $\bm{X} \sim \mathcal{N}(\bm{\mu}, \bm{\Sigma})$ as a Gaussian random vector, 
the expectation and variance of $\langle \bm{X},\bm{a}\rangle = \bm{X}\T \bm{a}$ are respectively $ \bm{\mu}\T \bm{a} $ and $\bm{a}\T\bm{\Sigma} \bm{a}$, or in other words, $\bm{X}\T \bm{a} \sim \mathcal{N}(\bm{\mu}\T \bm{a}, \bm{a}\T\bm{\Sigma} \bm{a})$.

\subsection{Gaussian Mixture Model}\ 

A Gaussian mixture model (GMM) in $\R^m$ is defined as the tuple $G=(\{p_j\}_j,\{\bm{\mu}_j\}_j, \{\bm{\Sigma}_j\}_j)$ 
where $j=1,2,\cdots, n$, $p_j \ge 0$ and $\sum_{j=1}^n p_j=1$, $\bm{\mu}_j \in \R^{m}$, and $\bm{\Sigma}_j \in M_m(\R)$ are positive-definite matrices. Under this notation, $n$ is called the component number, and the parameters $\{p_j\}_j,\{\bm{\mu}_j\}_j, \{\bm{\Sigma}_j\}_j$ are respectively called the mixing coefficients (fractionals), means, and covariances of the Gaussian components. 

The PDF of it $G$ is obtained by summing over all components:
$$\mathrm{PDF} (G)(\bm{x}) = \sum_{j=1}^n \frac{p_j}{(2\pi)^{m/2}\sqrt{\det(\bm{\Sigma_j})}} \exp\left(-\frac{(\bm{x}-\bm{\mu_j})\T \bm{\Sigma_j}^{-1} (\bm{x}-\bm{\mu_j})}{2}\right) $$

Here is another more understandable way of describing a Gaussian mixture model \cite{bishop2006pattern}.

Let $c$ be a categorical random variable of $n$ categories, with probability $p_j$ of being in the $j$-th category, i.e., $\P[c=j] = p_j$.
If $\bm{X} \sim G$, then the conditional distribution of $\bm{X}$ when $c=j$, denoted by $P(\bm{X} | c=j)$, is 
$$ P(\bm{X} | c=j) \sim \mathcal{N} (\bm{\mu}_j, \bm{\Sigma}_j) $$
The expectation of $\bm{X}$ is easily computed as $\E[\bm{X}] = \sum_{j=1}^n p_j \bm{\mu}_j$. The projection of $\bm{X}$ along unit vector $\bm{a}$ is also a random variable that follows a Gaussian mixture distribution, which is
$\langle \bm{X},\bm{a}\rangle = \bm{X}\T \bm{a} \sim \sum_{j=1}^{n} p_j \mathcal{N}(\bm{\mu}_j \T \bm{a}, \bm{a} \T \bm{\Sigma}_j \bm{a})$ with expectation $\mathbb{E}[\langle \bm{X}, \bm{a}\rangle]=\sum_{j=1}^{n} p_j \bm{\mu}_j \T \bm{a}$.

\subsection{The Expressiveness of GMMs}\ 

Although the Gaussian distribution is common in a variety of situations, there are some data distributions that differ significantly from the Gaussian distribution. 
Therefore, more expressive models are required to describe the real data distribution. 
In this part, the expressiveness of GMMs is characterized by the theorems below \cite{goodfellow2016deep,plataniotis2000gaussian}.

\begin{thm}
    Gaussian distributions are universal approximators, which can approximate any distribution by distribution. Namely, if $A$ is a distribution of a random variable $X$, then there exists a series of Gaussian mixture distributions, then there exists a series of GMMs $\{G_q\}_q \ (q \in \N)$ such that 
    $$ \{G_q\} \to A, \quad \text{by distribution.}$$
\end{thm}
\begin{proof}
    The proof can be found at page 6-7 of \cite{plataniotis2000gaussian}. 
\end{proof}

\begin{thm}
    Gaussian mixtures are uniquely identified by their distributions. If $G=(\{p_j\}_j,\{\bm{\mu}_j\}_j, \{\bm{\Sigma}_j\}_j)$ and $G'=(\{p'_k\}_k,\{\bm{\mu}'_k\}_k, \{\bm{\Sigma}'_k\}_k)$ are two GMMs with the same distribution, then their parameters are equal in the sense that they differ by one permutation. In other words, if $G$ and $G'$ are two GMMs with different set of parameters, then $G$ and $G'$ are distinguishable by distribution.
\end{thm}
\begin{proof}
    The proof can be found at page 7-8 of \cite{plataniotis2000gaussian}, or the Appendix of \cite{yan2023learning}.
\end{proof}

\subsection{Learning Gaussian Mixture Models}\ 

The commonly used methods of learning Gaussian mixtures can be roughly divided into two categories, namely iterative methods and gradient descent methods. Each method has its unique advantages and defects. Below is a list of some renowned methods for Gaussian mixture learning.

\subsubsection{The Expectation-Maximization and K-means Algorithm}\ 

The Expectation-Maximization (EM) algorithm and the K-means algorithm are iterative methods that iterate over the parameters of a GMM $G$ to fit $G$ to a distribution of data points, of which the EM algorithm is the most widely used. The classical EM algorithm contains 2 important steps, the Expectation (E) step and the Maximization (M) step. Each of these steps updates a part of the parameters. The two steps are performed alternatively until convergence is reached \cite{bishop2006pattern}.

The K-means algorithm is very similar to the Expectation-Maximization algorithm, except for that it uses \textit{hard assignments}, which means that every point is assigned to only one Gaussian component \cite{macqueen1967some}. 

However, these iteration-based approaches also have their drawbacks. 
For example, the Expectation-Maximization algorithm is known to suffer from the local optima problem. Under certain initializations, the EM algorithm might perform badly, converging to a bad local optima \cite{wu1983convergence}. 
Also, if the parameters of GMM $G$ are not explicitly given, such as the parameters are given by the output of the neural network, these methods will not work directly. 

\subsubsection{Gradient Descent Based Algorithms}\ 

There are a series of algorithms that fit GMMs by gradient descent. Generally speaking, the principal goal of gradient descent is to search for the optimal set of parameters $\bm{\theta}$ such that a certain loss function $L(\bm{\theta})$ attains its minimum. If $L$ is sufficiently differentiable, this is usually done by gradient descent (and its variations) over $L$. There are multiple gradient descent optimization algorithms, such as SGD, RMSProp or Adam, that achieve this goal in slightly different manners \cite{ruder2016overview}. 

However, the most crucial part is the designation of the loss function to be optimized. A good design of loss function is the key to successful learning or fitting of GMMs. 

One of the most commonly used loss functions, the \textit{Negative Log Likelihood (NLL) Loss} is defined as $L = -\log(H)$ where $H$ is the likelihood function. The term "negative log" comes directly form the formula. Since $-\log(x)$ is monotonically decreasing when $x>0$, minimizing the NLL loss is equivalent to maximizing the likelihood $H$. 
Given $ G=(\{p_j\}_j,\{\bm{\mu}_j\}_j, \{\bm{\Sigma}_j\}_j) (j=1,2, \cdots, n) , \ X = \{\bm{x}_i\}_i (i=1,2, \cdots, k)$, the likelihood $H$ is defined as follows:
$$H = \prod_{i=1}^{k} \left(\sum_{j=1}^{n} p_j \mathcal{N} (\bm{\mu}_j ,\bm{\Sigma}_j)(\bm{x}_i) \right) $$
Therefore, $L$ is obtained by 
$$L = -\log(H) = - \sum_{i=1}^k \log \left(\sum_{j=1}^{n} p_j \mathcal{N} (\bm{\mu}_j ,\bm{\Sigma}_j)(\bm{x}_i) \right) $$

There are also other gradient descent methods, such as the sliced Wasserstein distance \cite{kolouri2017sliced} or Wasserstein-Fischer-Rao gradient flow. \cite{yan2023learning}

Generally, some drawbacks of gradient descent for learning Gaussian mixture models are:

\begin{itemize}
\item \textbf{Local optima:} The loss functions may have multiple local maxima or minima. Gradient descent may get stuck in a poor solution that is not the global minimum. Till today, no loss function has theoretical guarantees to fit GMMs to global optima. To deal with this drawback, one may need to try multiple different initial values for the parameters or use some global optimization methods.

\item \textbf{Numerical instability:} Some loss functions suffer from heavy numerical instability. For example, the negative log likelihood loss for GMM computes the exponential function in the Gaussian density, which might cause overflow or underflow errors when the initialization is far from the data points, or when the covariance matrices are ill-conditioned.

\item \textbf{Slow convergence:} Generally speaking, gradient descent based methods is slower than iteration-based methods. To avoid missing the optima, the learning rate should be set small enough, therefore much more iterations are required to attain the optima. In addition, the gradient computation is another time-consuming step in gradient descent.
\end{itemize}

\section{Cram\'er Type Distances}\ 

Below we will introduce theoretical works about the Cram\'er type distances.

Note: Unless explicitly stated, we do not distinguish between a random variable and a probabilistic distribution in the following context, since those distances are defined solely over distributions, and each random variable has a distribution.

\subsection{The $L_\text{CDF}^p$ Class And The $l_p$-distance} \

Let $p \in [1, \infty)$ be a positive number. The $l_p$-distance \cite{bellemare2017cramer} between two probabilistic distributions $P, Q$ on $\R$ is defined as:
$$l_p(P, Q) = \left(\int_{-\infty}^{\infty} |\mathrm{CDF}(P) - \mathrm{CDF}(Q)|^p \dd x\right)^{1/p}$$
Where $\mathrm{CDF}$ denote the cumulative distribution function. 

Before we dive deeper into this section, we should check whether this distance is well-defined. The question is: on which space is the $l_p$-distance well-defined?

We know that a CDF function $F$ on $\R$ is right-continuous, non-decreasing with limit conditions 
$$ \lim_{x \to -\infty} F(x) = 0, \quad \lim_{x \to \infty} F(x) = 1 $$
We can write it as a set
$$\textbf{CDF} = \left\{F: \R \to [0,1] : F \text{ right continuous and non-decreasing}, \lim_{x \to -\infty} F(x) = 0,  \lim_{x \to \infty} F(x) = 1 \right\}$$
Since a CDF uniquely defines a distribution, we will not distinguish between a CDF and its corresponding distribution either, unless explicitly stated.

Let 
$$ H(x) = \left\{
\begin{aligned}
    & 0, &x<0 \\
    & 1, &x\ge 0
\end{aligned}
\right.$$
be the Heaviside function, which, according to the definitions above, is a CDF function. In fact, $H$ is the CDF of the degenerate distribution at $0$.

By now, we can define the function class $L_\text{CDF}^p$:
$$L_\text{CDF}^p = \{ F \in \textbf{CDF}: |F-H| \in L^p(\R) \} $$
Not all CDF functions belong to this class, though. Nonetheless, this is a sufficiently large class that contains the CDF of most distributions, including the Bernoulli distribution, the uniform distribution, and the Gaussian distribution.

We have the following lemma:
\begin{lemma} 
    \label{lemma:1}
    The space $(L_\text{CDF}^p, l_p)$ is a complete metric space that is closed under weighted average. In other words, it is a convex set.
\end{lemma}
For the proof, please see Appendix \ref{appendix:a}.

The $l_p$-distance, especially for $p=2$, has many intriguing properties. When $p=2$, the distance is called Cram\'er 2-distance, denoted by $C_2$. 
It has unbiased sample gradient and contraction property \cite{bellemare2017cramer}.

In the following, we will mainly focus on the Cram\'er 2-distance of Gaussian distributions and Gaussian mixtures. 

From now on, we denote the cumulative distribution function of the standard normal distribution by 
$$\Phi(x) = \int_{-\infty}^{x} \frac{\exp\left(-\frac{y^2}{2}\right)}{\sqrt{2\pi}} \dd y$$
Then we define the cumulative distribution function $\Phi_{\mu, \sigma^2}$ of normal distribution $\mathcal{N}_{\mu, \sigma^2}$: $\Phi_{\mu, \sigma^2}(x) := \Phi((x-\mu)/\sigma)$, and $\Phi_{\mu, \sigma^2}^{\text{c}} (x) := 1-\Phi((x-\mu)/\sigma)$. By definition, $\Phi_{0,1} (x) = \Phi (x)$.

The following lemma might be useful:
\begin{lemma}
    \label{lemma:2}
    GMMs are dense in $L_\text{CDF}^2$.
\end{lemma}
See Appendix \ref{appendix:a} for the proof.

\subsection{A Heuristic Computation}\ 

Suppose that we want to compute the Cram\'er $2$-distance between two Gaussian distributions: $\mathcal{N}_{m, s^2}$ and $\mathcal{N}_{0, 1}$.
We have 
$$
\begin{aligned}
\int_{-\infty}^{\infty} |\Phi_{m,s^2}(x)-\Phi(x)|^2 \dd x 
&= \int_{-\infty}^{\infty} \Phi_{m,s^2}(x)(1-\Phi(x)) \dd x + \int_{-\infty}^{\infty} (1-\Phi_{m,s^2}(x)) \Phi(x) \dd x\\
&  + \int_{-\infty}^{\infty} \Phi(x)(1-\Phi(x)) \dd x + \int_{-\infty}^{\infty} \Phi_{m,s^2}(x)(1-\Phi_{m,s^2}(x)) \dd x\\
\end{aligned}
$$
For simplicity, we only compute this term $\int_{-\infty}^{\infty} (1-\Phi_{m,s^2}(x)) \Phi(x) \dd x$,
which provides us enough information to derive the other 3 terms by analogy.

We take derivative of $m$ twice:

$$
\begin{aligned}
\frac{\partial^2}{\partial m^2} \int_{-\infty}^{\infty} (1-\Phi_{m,s^2}(x)) \Phi(x) \dd x
&= \int_{-\infty}^{\infty} \frac{\partial^2}{\partial m^2}\left(1-\Phi\left(\frac{x-m}{s}\right)\right) \Phi(x) \dd x\\
&= \int_{-\infty}^{\infty} \frac 1 s \frac{\partial}{\partial m} \Phi'\left(\frac{x-m}{s}\right) \Phi(x) \dd x\\
&= \int_{-\infty}^{\infty} -\frac 1 {s^2} \Phi''\left(\frac{x-m}{s}\right) \Phi(x) \dd x\\
&= \int_{-\infty}^{\infty} \frac 1 {s} \Phi'\left(\frac{x-m}{s}\right) \Phi'(x) \dd x & \text{(Integration by parts)} \\
&= \frac{1}{2\pi s}\int _{-\infty} ^{\infty} \exp\left(-\frac{
(s^2+1) \left(x+\frac{m s^2}{s^2+1}\right)^2 + \frac{m ^2s^2}{s^2+1}}{2s^2} \right) \dd x \\
&= \frac{\sqrt{\frac{2\pi s^2}{s^2+1}}}{2\pi s} \exp\left(-\frac{m ^2}{2(s^2+1)}\right)\\
&= \frac{1}{\sqrt{2\pi(s^2+1)}} \exp\left({-\frac{m ^2}{2(s^2+1)}}\right)\\
\end{aligned}
$$
Integrate $m$ back:
$$
\frac{\partial}{\partial m} \int_{-\infty}^{\infty} (1-\Phi_{m,s^2}(x)) \Phi(x) \dd x = \Phi_{0, s^2+1}(m) +C
$$
Where $ C = 0$ by taking the limit at $ m \to -\infty$. Integrate again:
$$
\int_{-\infty}^{\infty} (1-\Phi_{m,s^2}(x)) \Phi(x) \dd x = \Phi_{0, s^2+1}^{(-1)}(m) + C_1 = \sqrt{s^2+1} \cdot \Phi^{(-1)}\left(\frac{m}{\sqrt{s^2+1}}\right) + C_1
$$
Where $\Phi^{(-1)}$ denote the antiderivative of $\Phi$.

It's easy to verify (although may not be known to all) by integration by parts that 
$$
\Phi^{(-1)}(x) = x\Phi(x) + \frac{1}{\sqrt{2\pi}}\exp\left(-\frac{x^2}{2}\right) + C_0
$$
In our case,  $C_0 = C_1 = 0$ by taking $m \to -\infty$. In conclusion, 
$$
\int_{-\infty}^{\infty} (1-\Phi_{m,s^2}(x)) \Phi(x) \dd x = \sqrt{s^2+1} \cdot U \left(\frac{m}{\sqrt{s^2+1}}\right)
$$
Where 
$$U(x) = x\Phi(x) + \frac{1}{\sqrt{2\pi}}\exp\left(-\frac{x^2}{2}\right) = \mathrm{GELU}(x) + \frac{1}{\sqrt{2\pi}}\exp\left(-\frac{x^2}{2}\right)$$
Here, $\mathrm{GELU}(x) = x\Phi(x)$ means the Gaussian Error Linear Unit function \cite{hendrycks2016gelu}. Note that the function $U(x)$ here is exactly the anti-derivative of the function $\Phi(x)$, i.e., $U'(x) = \Phi(x)$.

Then, we can compute the integral $\int_{-\infty}^{\infty} (1-\Phi_{m_1,s_1^2}(x)) \Phi_{m_2,s_2^2}(x) \dd x$ by changing of variables:
$$
\begin{aligned}
\int_{-\infty}^{\infty} (1-\Phi_{m_1,s_1^2}(x)) \Phi_{m_2,s_2^2}(x) \dd x &= \int_{-\infty}^{\infty} (1-\Phi_{m_1-m_2,s_1^2}(y)) \Phi_{0,s_2^2}(y) \dd y \\
&= s_2 \int_{-\infty}^{\infty} (1-\Phi_{(m_1-m_2)/s_2,s_1^2/s_2^2}(y)) \Phi_{0,1}(y) \dd y \\
&= s_2 \sqrt{\frac{s_1^2}{s_2^2} +1} \cdot U \left(\frac{\frac{m_1-m_2}{s_2}}{\sqrt{\frac{s_1^2}{s_2^2} +1}}\right) \\
&= \sqrt{s_1^2+s_2^2} \cdot U \left(\frac{m_1-m_2}{\sqrt{s_1^2+s_2^2}}\right)
\end{aligned}
$$
For $s_2 = 0$, we can just take the limit
$$\lim_{s_2 \to 0} \sqrt{s_1^2+s_2^2} \cdot U \left(\frac{m_1-m_2}{\sqrt{s_1^2+s_2^2}}\right) = s_1 \cdot U \left(\frac{m_1-m_2}{s_1}\right)$$

\subsection{The Closed Formula for Cram\'er 2-Distance of 1D GMMs}\ 

The main work of this article is the full parametric form expression for the Cram\'er 2-distance of two univariate Gaussian mixtures.
This function is of central importance in this study and is used multiple times in
subsequent analysis and experiments.

Consider 2 univariate Gaussian mixture distributions $G_1=(\{p_j\}_j,\ \{\mu_j\}_j,\ \{\sigma_j^2\}_j)\ (j=1,2, \cdots, n)$ and $G_2=(\{p_k'\}_k,\ \{\mu_k'\}_k,\ \{\sigma_k'^2\}_k)\ (k = 1,2,\cdots, n')$. The Cram\'er 2-distance is defined as 
$$C_2(G_1, G_2) = \left(\int_{-\infty}^{\infty} |\mathrm{CDF}(G_1)(x) - \mathrm{CDF}(G_2)(x)|^2 \dd x\right)^{1/2}$$
The CDF (cumulative distribution function) of $G_1$ and $G_2$ are separately:
$$
\begin{aligned}
    \mathrm{CDF}(G_1)(x) &= \sum _{j=1}^n p_j \Phi_{\mu_j, \sigma_j^2}(x) \\
    \mathrm{CDF}(G_2)(x) &= \sum _{k=1}^{n'} p'_k \Phi_{\mu'_k, \sigma_k'^2}(x) \\
\end{aligned}
$$
Now we can derive the formula
$$
\begin{aligned}
    C_2^2(G_1, G_2) &= \int_{-\infty}^{\infty} |\mathrm{CDF}(G_1)(x) - \mathrm{CDF}(G_2)(x)|^2 \dd x \\
    &= \int_{-\infty}^{\infty} (\mathrm{CDF}(G_1)(x) - \mathrm{CDF}(G_2)(x)) ((1-\mathrm{CDF}(G_2)(x)) - (1-\mathrm{CDF}(G_1)(x))) \dd x\\
    &= \int_{-\infty}^{\infty} \mathrm{CDF}(G_1)(x) (1-\mathrm{CDF}(G_2)(x)) \dd x + \int_{-\infty}^{\infty} \mathrm{CDF}(G_2)(x) (1-\mathrm{CDF}(G_1)(x)) \dd x \\
    &- \int_{-\infty}^{\infty} \mathrm{CDF}(G_1)(x) (1-\mathrm{CDF}(G_1)(x)) \dd x - \int_{-\infty}^{\infty} \mathrm{CDF}(G_2)(x) (1-\mathrm{CDF}(G_2)(x)) \dd x \\
    &= \int_{-\infty}^{\infty} \sum _{j=1}^n \sum _{k=1}^{n'}  \left(p_j \Phi_{\mu_j, \sigma_j^2}(x)  p'_k \Phi^{\text{c}}_{\mu'_k, \sigma_k'^2}(x) \right) \dd x + \cdots & \text{(By analogy)}\\
    &= \sum _{j=1}^n \sum _{k=1}^{n'} \left( p_j p_k' \sqrt{\sigma_j^2+\sigma_k'^2} \cdot  U \left(\frac{\mu_j-\mu_k'}{\sqrt{\sigma_j^2+\sigma_k'^2}}\right) \right) + \cdots\\
\end{aligned}
$$
We write the full formula below in case someone fails on the analogy:
\begin{equation}
    \label{eqn:c2}
    \begin{aligned}
        C_2^2(G_1, G_2) &= \sum _{j=1}^n \sum _{k=1}^{n'} \left( p_j p_k' \sqrt{\sigma_j^2+\sigma_k'^2} \cdot  U \left(\frac{\mu_j-\mu_k'}{\sqrt{\sigma_j^2+\sigma_k'^2}}\right) \right)
        + \sum _{j=1}^n \sum _{k=1}^{n'} \left( p_j p_k' \sqrt{\sigma_j^2+\sigma_k'^2} \cdot  U \left(\frac{\mu_k'-\mu_j}{\sqrt{\sigma_j^2+\sigma_k'^2}}\right) \right) \\
        &- \sum _{j=1}^n \sum _{k=1}^n \left( p_j p_k \sqrt{\sigma_j^2+\sigma_k^2} \cdot  U \left(\frac{\mu_j-\mu_k}{\sqrt{\sigma_j^2+\sigma_k^2}}\right) \right) 
        - \sum _{j=1}^{n'} \sum _{k=1}^{n'} \left( p_j' p_k' \sqrt{\sigma_j'^2+\sigma_k'^2} \cdot  U \left(\frac{\mu_j'-\mu_k'}{\sqrt{\sigma_j'^2+\sigma_k'^2}}\right) \right) \\
    \end{aligned}
\end{equation}
In fact, we have a more symmetric form. If we denote $V(x) = (U(x)+U(-x))/2$ for $x \in \R$, we have
\begin{equation}
    \label{eqn:c2other}
    \begin{aligned}
        C_2^2(G_1, G_2) &= 2\sum _{j=1}^n \sum _{k=1}^{n'} \left( p_j p_k' \sqrt{\sigma_j^2+\sigma_k'^2} \cdot  V \left(\frac{\mu_j-\mu_k'}{\sqrt{\sigma_j^2+\sigma_k'^2}}\right) \right) \\
        &- \sum _{j=1}^n \sum _{k=1}^n \left( p_j p_k \sqrt{\sigma_j^2+\sigma_k^2} \cdot  V \left(\frac{\mu_j-\mu_k}{\sqrt{\sigma_j^2+\sigma_k^2}}\right) \right) 
        - \sum _{j=1}^{n'} \sum _{k=1}^{n'} \left( p_j' p_k' \sqrt{\sigma_j'^2+\sigma_k'^2} \cdot  V \left(\frac{\mu_j'-\mu_k'}{\sqrt{\sigma_j'^2+\sigma_k'^2}}\right) \right) \\
    \end{aligned}
\end{equation}
which saves about $1/4$ of the computation.

Although the functions $U$ and $V$ are not elementary functions (the Gaussian error linear unit function itself is not elementary), it is provided by common machine learning libraries such as PyTorch \cite{pytorchgelu}. So it is a good idea to directly implement such a function and to directly perform gradient descent over it. The example implementation can be found in the Appendix \ref{appendix:b}.

The following theorem ensures the gradient stability of Cram\'er 2-distance:
\begin{thm}
    \label{thm:bounded}
    Suppose that $G_1=(\{p_j\}_j,\ \{\mu_j\}_j,\ \{\sigma_j^2\}_j)\ (j=1,2, \cdots, n)$ is the online distribution to be trained, and $G_2=(\{p_k'\}_k,\ \{\mu_k'\}_k,\ \{\sigma_k'^2\}_k)\ (k = 1,2,\cdots, n')$ is the target distribution. The loss function is $L=C_2^2(G_1, G_2)$. Then for any $j=1,2, \cdots, n$, we have
    $$ \left| \frac {\partial L} {\partial \mu_j} \right| \le 4, \quad \left| \frac {\partial L} {\partial \sigma_j} \right| \le 4.$$ 
    In other words, loss $L$ is global Lipschitz for $\{\mu_j\}$ and $\{\sigma_j\}$.
\end{thm}
The proof can be found in the Appendix \ref{appendix:a}.

Remark: The GELU function has a well known approximate form \cite{pytorchgelu}
$$\mathrm{GELU}(x) \approx \frac x 2 \left(1+ \tanh \left(\sqrt{\frac 2 \pi} \left(x + 0.044715x^3\right)\right)\right)$$
We do not use this form in any of our experiments, because we want an accurate computation of the loss values and gradients.

\subsection{Sliced Cram\'er 2-Distance for the Multivariate Case}\ 

This section is a natural generalization of the formula in the univariate case, similar to \cite{kolouri2017sliced} and \cite{kolouri2020sliced}.

Let $\bm{X}$ and $\bm{Y}$ be random vectors in $\R^m$. The Sliced Cram\'er 2-distance (also called the Cram\'er-Wold distance) for $X$ and $Y$ could be defined as follows:
$$S_2^2(\bm{X}, \bm{Y}) := \int_{\bm{\nu} \in \mathbb{S}^{m-1}} C_2^2(\langle \bm{X}, \bm{\nu}\rangle ,  \langle \bm{Y}, \bm{\nu}\rangle) \dd \bm{\nu} $$
where $\langle \_ , \bm{\nu} \rangle$ denote the projection onto the direction of $\bm{\nu}$. 

For simplicity of calculation, We uniformly and independently sample $t$ unit vectors $\{\bm{\nu}_i\} (i=1, 2, \cdots, t)$ from the sphere $\mathbb{S}^{m-1} \subset \R^m$. Then we approximate $S_2$ by 
$$S_2^2(\bm{X}, \bm{Y}) \approx \sum_{i=1}^{t} C_2^2(\langle \bm{X}, \bm{\nu}_i\rangle ,  \langle \bm{Y}, \bm{\nu}_i\rangle)$$
Note that if $\bm{X} \sim G = (\{p_j\}_j,\{\bm{\mu}_j\}_j, \{\bm{\Sigma}_j\}_j) (j=1,2, \cdots, n) $ is a multivariate GMM, then $\langle \bm{X}, \bm{\nu}\rangle$ yields a univariate GMM by projection onto the direction of unit vector $\bm{\nu}$: 
$$\langle \bm{X}, \bm{\nu}\rangle \sim G_{\bm{\nu}} = \left(\{p_j\}_j, \{\bm{\mu}_j\T\bm{\nu}\}_j, \{\bm{\nu} \T \bm{\Sigma}_j \bm{\nu}\}_j\right).$$

Here is a figure that demonstrates how this formula works.

\begin{center}
    \includegraphics[width=0.95\columnwidth]{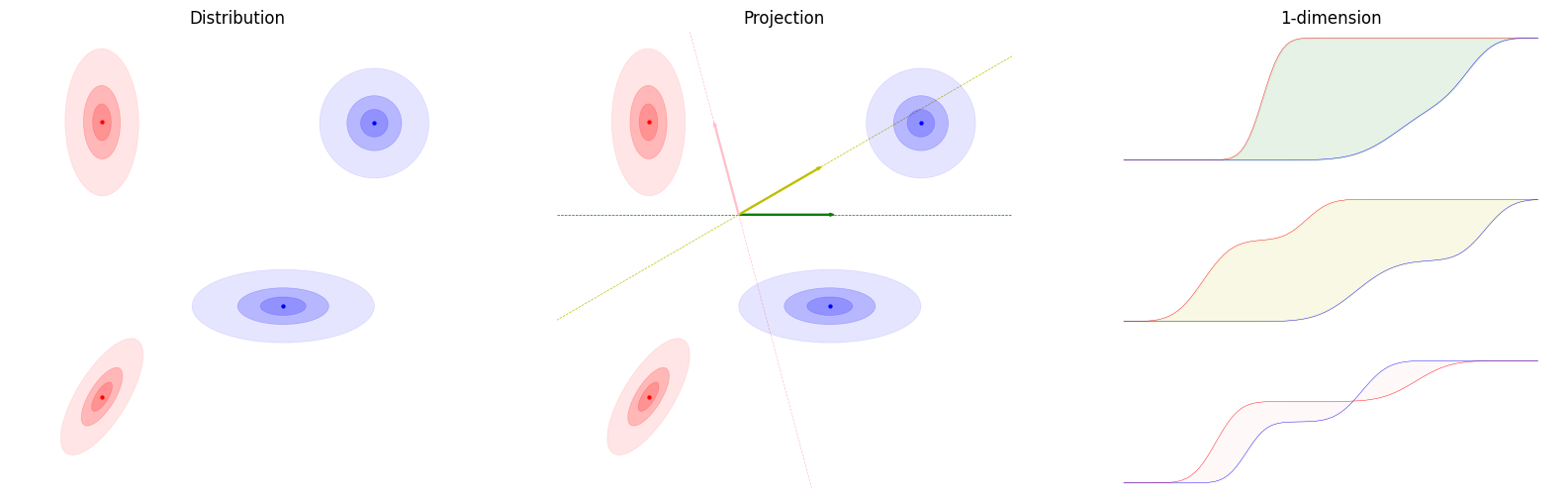} 
    \captionof{figure}{Demonstration of sliced Cram\'er 2-distance.}
\end{center}

Again, we confirm that this is a well-defined distance.
\begin{thm}
    \label{thm:sc2dist}
    The function
    $$S_2(\bm{X}, \bm{Y}) = \sqrt{\int_{\bm{\nu} \in \mathbb{S}^{m-1}} C_2^2(\langle \bm{X}, \bm{\nu}\rangle ,  \langle \bm{Y}, \bm{\nu}\rangle) \dd \bm{\nu}}$$
    defines a distance of two distributions.
\end{thm}

\begin{proof}
    The proof of symmetry and triangle inequality is direct. To prove the positivity, one need to show that 
    $$\mathrm{CDF} (\langle \bm{X}, \bm{\nu}\rangle) = \mathrm{CDF} (\langle \bm{Y}, \bm{\nu}\rangle) \ (\forall \bm{\nu}) \implies \bm{X} \sim \bm{Y} .$$
    The left side implies $$\langle \bm{X}, \bm{\nu}\rangle \sim \langle \bm{Y}, \bm{\nu}\rangle \ (\forall \bm{\nu}), \text{  (as distribution) } $$
    Which is the Cram\'er-Wold Theorem \cite{cramer1936some,billingsley1995probability}, and can be proved by the fact that Radon transform admits an inverse.
\end{proof}

We show that Sliced Cram\'er 2-distance inherits some key properties from the univariate Cram\'er 2-distance. These results apply in a general sense, not just GMMs. 
\begin{thm}
    \label{thm:sc2prop}
    Sliced Cram\'er 2-loss enjoys the following properties in general:
    \begin{itemize}
        \item Independent sum: For two random vectors $\bm{X}$, $\bm{Y}$, and a random vector $\bm{A}$ independent of both  $\bm{X}$ and $\bm{Y}$. Then 
        $$S_2^2(\bm{A}+\bm{X}, \bm{A}+\bm{Y}) \le S_2^2(\bm{X}, \bm{Y})$$
        \item Scaling property: For two random vectors $\bm{X}$, $\bm{Y}$, and $c>0$,  $$S_2^2(c\bm{X}, c\bm{Y}) =c S_2^2(\bm{X}, \bm{Y})$$
        \item Unbiased sampling gradients: Given $\mathcal{X} = \bm{X}_1, \cdots, \bm{X}_r$ sampled from a distribution $P$, the empirical distribution $ \hat{P} = \frac 1 r (\delta_{\bm{X}_1}+ \cdots + \delta_{\bm{X}_r})$, and a distribution $G_\theta$ induced by parameter $\theta$,
        $$\E _{\mathcal{X}\sim P} \left[\nabla_\theta S_2^2 (G_\theta, \hat{P})\right] = \nabla_\theta S_2^2 (G_\theta, P)$$
        Moreover, If $\bm{\nu}$ is a random unit vector uniformly distributed in $\mathbb{S}^{m-1}$, we have
        $$B_{m-1}\cdot \E _{\bm{\nu}} \E _{\mathcal{X}\sim P} \left[\nabla_\theta C_2^2 (\langle G_\theta, \bm{\nu}\rangle, \langle \hat{P}, \bm{\nu}\rangle)\right] = \nabla_\theta S_2^2 (G_\theta, P)$$
        Where $B_{m-1} = 2\pi^{m/2}/\Gamma(m/2)$ is the hypersurface area of $\mathbb{S}^{m-1}$. 
    \end{itemize} 
\end{thm}

Just like the univariate case, we have the gradient boundedness theorem for Sliced Cram\'er 2-loss for multivariate GMMs as well:
\begin{thm}
    \label{thm:boundedsc2}
    Suppose that $G_1=(\{p_j\}_j,\ \{\bm{\mu}_j\}_j,\ \{\bm{\Sigma_j}\}_j)\ (j=1,2, \cdots, n)$ is the online distribution to be trained, and $G_2=(\{p_k'\}_k,\ \{\bm{\mu}_k'\}_k,\ \{\bm{\Sigma}_k'\}_k)\ (k = 1,2,\cdots, n')$ is the target distribution. The loss function $L=S_2^2(G_1, G_2)$. Then for any $j=1,2, \cdots, n$, we have
    $$ \left| \nabla_{\bm{\mu}_j} L \right| \le 4B_{m-1}$$
    and if we obtain $\bm{\Sigma_j}$ by $\bm{S}_j\T \bm{S}_j$ where $\bm{S}_j$ is a learnable matrix, then
    $$ \quad \left| \nabla_{\bm{S}_j} L \right| \le 4B_{m-1}$$
    Where $B_{m-1} = 2\pi^{m/2}/\Gamma(m/2)$.
\end{thm}

Although we have tried to derive a full parametric form for a distance of general multivariate GMMs, we have simply failed because of the intrinsic complexity of the formula. Yet, our approaches still offer unbiased gradient guarantees, anisotropic Gaussian support, and simpler implementation compared to \cite{smieja2020segma}.

\section{Experiments and Results}\ 

In order to demonstrate the feasibility and effectiveness of learning GMMs by gradient descent over (Sliced) Cram\'er 2-distance, we have conducted experiments for both the univariate and the multivariate case.

\subsection{Distributional Q-Learning}\

Distributional Q-Learning \cite{sutton2018reinforcement,bellemare2017distributional,bellemare2023distributional} 
is a model-free reinforcement learning algorithm which learns the distribution of the returns given a state-action pair, rather than only the expectation of outcome. 
If we denote $(s,a)$ by the state-action pair, $R(s,a)$ the reward over $(s,a)$, $(S',A')$ be the subsequent state-action pair, and $Z$ the distribution of returns, then the Bellman Operator can be written as
$$Z(s,a) \gets R(s,a) + \gamma Z(S', A') \text{ (as distribution)}$$

Distributional returns contain more information than scalar returns, including the expectations, variances, momentums and risks. This allows the agent to capture the risk preferences of the policy, thus can improve the stability and performance of deep neural network agents.

Here are some famous examples of distributional Q-learning:
\begin{itemize}
    \item \textbf{C51} (Categorical 51) \cite{bellemare2017distributional}: This method discretizes the return distribution into 51 equally spaced atoms (deltas) at fixed points on the interval $[-10,10]$, and learns a categorical distribution over them. 
    It uses a projection operator to update the distribution parameters based on the Bellman equation, and greatly outperforms DQN on the Atari57 benchmark. 
    \item \textbf{QR-DQN} (Quantile Regression - Deep Q Network) \cite{dabney2018qr}: This method discretizes the return distribution into $N$ atoms with fixed probabilities but adjustable positions (called quantiles), and it improved further upon C51.
    \item \textbf{FQF} (Fully Quantile Function) \cite{yang2020fully}: This method discretizes the return distribution into $N$ atoms with both adjustable probabilities (given by a fractional proposal network) and adjustable positions. The parameters are updated by 1-Wasserstein distance. FQF improved even further upon QR-DQN.
\end{itemize}
All these methods use a mixture of delta (degenerate) distributions, of which the CDF are not continuous and show "zig-zags" in their plots. However, considering the expressiveness of GMMs, it's entirely possible to learn a mixture of Gaussians towards the distribution. Given the continuity and smoothness of the CDF, Such a model could be capable of capturing fine-grained details of the distribution in fewer parameters.

It's worth noting that we are not the first one to propose such an idea. In the article \cite{choi2019distributional}, a Gaussian mixture deep Q network is learned, but the loss function used is \textit{Jensen-Tsallis Distance}, which is the $L^2$ difference of two \textit{probability density functions (PDF)}, not \textit{cumulative distribution functions (CDF)}. 
We are also not the first to apply the Cram\'er distance to distributional reinforcement learning. The Cram\'er distance have been successfully tested on a Quantile Regression DQN, which improves over the original QR-DQN \cite{lheritier2022cramer}.
But by now, thanks to the formula of the Cram\'er 2-distance between two GMMs earlier, it is now feasible to combine the two techniques together, yielding a prosperous architecture.

To test the effectiveness, we designed a distributional DQN, a simple 3-layer full-connection network. The input size is the observation space, with 2 hidden layers of size 128, and output 3 parts: fractional $\{p_j\}$, mean $\{\mu_j\}$ and standard deviation $\{\sigma_j\}$. The total output dimension is \texttt{3 * Number\_of\_mixtures * Action\_dimension}. The network architecture is the same to \cite{choi2019distributional}, but the loss function is our own. 
Without enough computational resources, we only tested the Gymnasium \texttt{LunarLander-v2} \cite{gymnasiumlunar}. 
This is because this environment possesses some intrinsic randomness, such as the shape of the terrain. Some hyperparameters are listed in this table:

\begin{center}
\begin{tabular}{cc|cc}
\hline 
Parameter & Value & Parameter & Value \\
\hline 
Hidden layer count & 2 & Hidden layer size & 128\\
Discount rate ($\gamma$) & 0.99 & Number of mixtures & 3 \\
Observation dimension & 8 & Action dimension & 4 \\
Batch size & 64 & Target update in frames & 200 \\
Main learning rate & 5e-5 & Fractional proposal part learning rate & 5e-9\\
Optimizer & Lion \cite{chen2023symbolic} & Replay capacity & 1e+5\\
\end{tabular}
\captionof{table}{Hyperparameters}
\end{center}

We use the Double DQN \cite{van2016deep}
which consists of an online network for training and action selection, and a target network for the estimation of Q value distribution.
The main motivation is that Double DQN is a practical solution in order to address overestimation of the mean $\{\mu_j\}$ and standard deviation $\{\sigma_j\}$ parts with little costs.
Note that the network of parameter $\theta$ returns a univariate Gaussian mixture distribution $Z_\theta(S,A)$. Therefore, the loss function (Double DQN) can be written as:
$$L = \frac{1}{\text{batch\_size}} \sum_{(S,A,R,S')\in \text{Batch}} C_2^2\left(Z_{\theta_\text{online}}(S,A), \ R+\gamma Z_{\theta_\text{target}}\left(S',\arg \max_{a \in \mathcal{A}} \mathbb{E}[Z_{\theta_\text{online}}(S', a)]\right)\right)$$

The algorithm is shown as follows.
\begin{algorithm}
    \caption{Computation of Cram\'er 2-loss of GMM DQN (Double DQN version)}
    \begin{algorithmic}[1]
        \Procedure{Cram\'er2loss} {}
        \State Randomly sample a batch of $(S,A,R,S')$ from the replay memory
        \State $L \gets 0$
        \ForAll {$(S,A,R,S')$ in batch}
            \State // Input distribution
            \State $(\{p_j\}, \{\mu_j\}, \{\sigma_j\}) \gets Z_{\theta_\text{online}}(S,A)$
            \State // Selection of action
            \ForAll {$a$ in the action set}
                \State $(\{p_{a,j}\}, \{\mu_{a,j}\}, \{\sigma_{a,j}\}) \gets Z_{\theta_\text{online}}(S',a)$
                \State $ q_a \gets \sum_{j=1}^n p_{a,j} \mu_{a,j}$ 
            \EndFor
            \State $ a_0 \gets \arg \max_{a \in \mathcal{A}} q_a$ 
            \State // Target distribution
            \State $(\{p'_j\}, \{\mu'_j\}, \{\sigma'_j\}) \gets Z_{\theta_\text{target}}(S',a_0)$
            \For {$j=1$ to $n$}
                \State $\mu'_j \gets R + \gamma \mu'_j$
                \State $\sigma'_j \gets \gamma \sigma'_j$
            \EndFor
            \State // Compute loss according to the previous formula
            \State $L \gets L + C_2^2\left((\{p_j\}, \{\mu_j\}, \{\sigma_j\}),\ \  (\{p'_j\}, \{\mu'_j\}, \{\sigma'_j\})\right)$
        \EndFor
        \State {$L \gets (L / \text{batch\_size})$} 
        \State
        \Return $L$
        \EndProcedure
    \end{algorithmic}
\end{algorithm}
The rest of the training procedure is the same as Double DQN. 

Another important factor to consider is the restrictions on $\{p_j\}$ and $\{\sigma_j\}$ parts.
We use a Softmax function to obtain the fractional part $\{p_j\}$, and set a small learning rate (5e-9) for this part to avoid it from degenerating. 
For the standard deviation part $\{\sigma_j\}$, we should prevent them from being negative, which lose their mathematical meanings and affect both performance and interpretability. 
In our experiments, this is done by adding a large penalty term over negative parts of $\{\sigma_j\}$:
$$ L \gets L + 10 \sum_{j=1}^n \mathrm{ReLU}(-\sigma_j)$$
The coefficient $10$ is enough, due to our previous theorem \ref{thm:bounded}.

We achieved a score of $279 \pm 22$ in \texttt{LunarLander-v2}. The figures below illustrate the behavior of the agent and the corresponding distributions in a $313$-point perfect landing. 
\begin{center}
\begin{tabular}{c}
    \includegraphics[width=1.0\columnwidth]{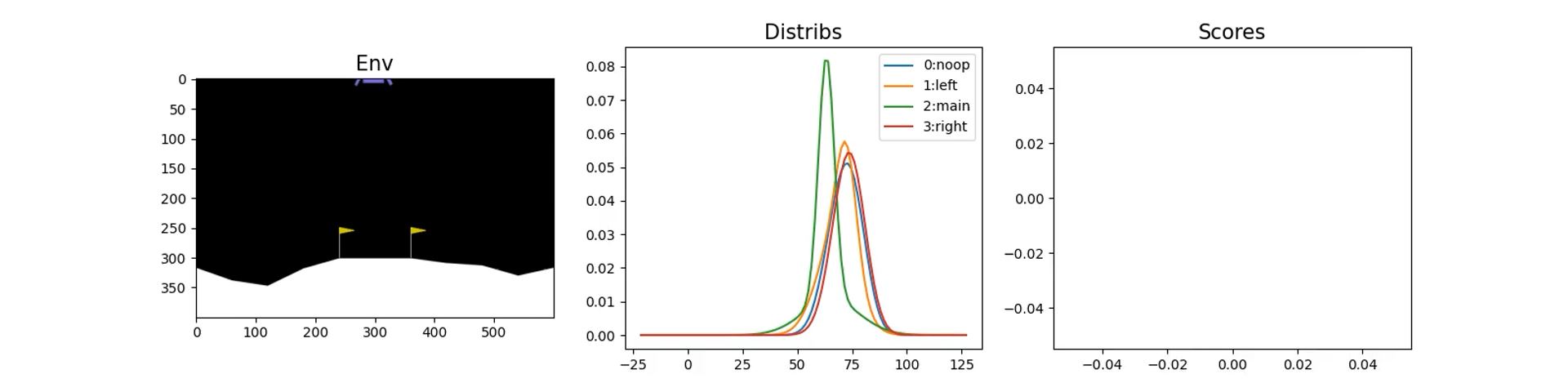} \\
    \includegraphics[width=1.0\columnwidth]{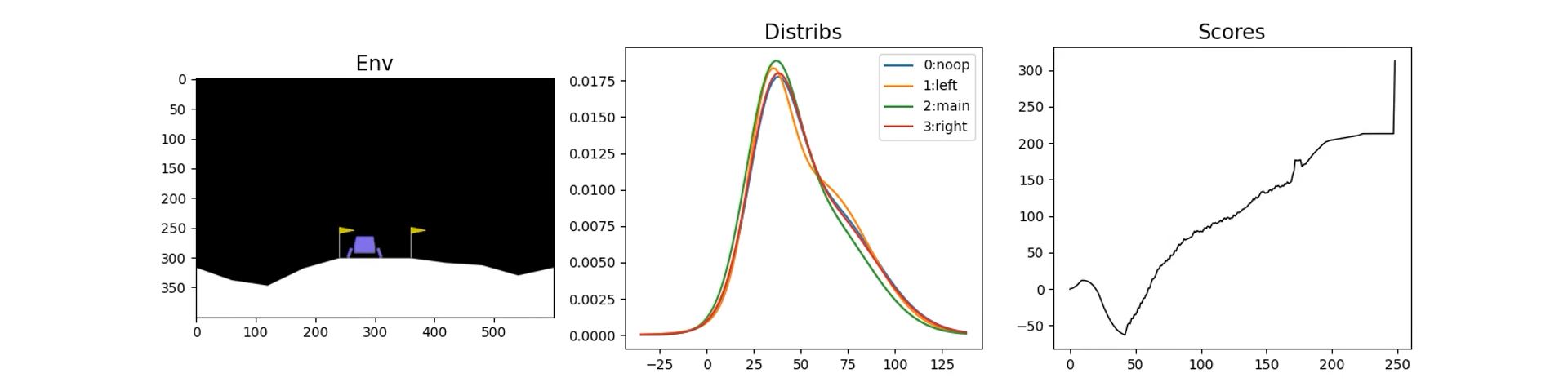} \\
\end{tabular}
\captionof{figure}{DQN experiment results.}
\end{center}
The result shows that the agent is able to learn complex distributions as well as evaluating and distingushing between different actions. 

\subsection{Multivariate GMM Learning}\ 

From our earlier discussions on the Sliced Cram\'er 2-distance, it is theoretically feasible to learn a general multivariate GMM towards another target GMM. Specifically, a set of $n'$ data points can be considered as the mixture of $n'$ degenerate Gaussians. The algorithm, especially the procedure of loss computation are shown in the following pseudo-code:
\begin{algorithm}
    \caption{Computation of Sliced Cram\'er 2-loss of multivariate GMMs}
    \begin{algorithmic}[1]
        \Procedure{SlicedCram\'er2loss} {}
        \State Input GMM: $G=(\{p_j\}_j,\{\bm{\mu}_j\}_j, \{\bm{\Sigma}_j\}_j)\  (j=1,2, \cdots, n)$
        \State Target GMM: $ G'=(\{p'_k\}_k,\{\bm{\mu}'_k\}_k, \{\bm{\Sigma}'_k\}_k)\  (k=1,2, \cdots, n') $
        \State // If we fit a GMM towards a set of $n'$ points, then $ G'=(\{1/n'\}_k,\{\bm{x}_k\}_k, \{\bm{0}\}_k)$
        \State Number of projections (slices): $t$
        \State Uniformly sample $\bm{\nu}_1, \bm{\nu}_2, \cdots \bm{\nu}_t \in \mathbb{S}^{m-1}$
        \State $L \gets 0$
        \For {$i=1$ to $t$}
            \State // projection onto $\bm{\nu}_i$
            \State $G_{\bm{\nu}_i} \gets \left(\{p_j\}_j,\{\bm{\mu}_j\T \bm{\nu}_i\}_j, \{\bm{\nu}_i\T \bm{\Sigma}_j \bm{\nu}_i\}_j\right)$
            \State $G'_{\bm{\nu}_i} \gets \left(\{p'_k\}_k,\{{(\bm{\mu}'_k)} \T \bm{\nu}_i\}_k, \{\bm{\nu}_i\T \bm{\Sigma}'_k \bm{\nu}_i\}_k\right)$
            \State $L \gets L + S_2^2(G_{\bm{\nu}_i},G'_{\bm{\nu}_i})$
        \EndFor
        \Return $L$
        \EndProcedure
    \end{algorithmic}
\end{algorithm}

To demonstrate its feasibility, we fit a multivariate GMM to a fixed data distribution, using the algorithm above. We tested it on a small dataset (which is the same dataset in \cite{kolouri2017sliced}, available at GitHub repository \cite{kolouriswgmmrepo}) 
with 850 points ($n'=850$) on a plane (dimension $m=2$), forming a rectangle, a circle, and a line attached to them. The GMM contains 10 mixtures ($n=10$). We ran this experiment across 3 different random seeds: 123, 456 and 789.

For $G=(\{p_j\}_j,\{\bm{\mu}_j\}_j, \{\bm{\Sigma}_j\}_j)$, considering the restrictions on them, we obtain them separately with different learning rates as follows: 
\begin{itemize}
    \item Fractional part $\{p_j\}_j$: By applying a Softmax function to $n$ parameters, we obtain an $n$-category distribution. The learning rate for this part is set to 5e-6. We set small learning rate for this part in order to prevent it from degenerating. 
    \item Mean part $\{\bm{\mu}_j\}$: This part is learned directly as $n$ $m$-dimensional vectors. The learning rate for this part is set to 2e-2.
    \item Covariance part $\{\bm{\Sigma}_j\}_j$: By $\bm{\Sigma}_j = \bm{S}_j\T \bm{S}_j$ where $\bm{S}_j \in M_m(\R)$ is the learnable matrix, in order to ensure the positive-definiteness. The learning rate for this part is set to 3e-3.
\end{itemize}
Again, we use the Lion (Evo\textbf{l}ved S\textbf{i}gn M\textbf{o}me\textbf{n}tum) optimizer \cite{chen2023symbolic} because it is easy to understand and implement.

Note: In our experiment, the dimension $m=2$. Due to the particular shape of $\mathbb{S}^1$ (which is a circle), we are able to equidistantly sample $\bm{\nu}_1, \bm{\nu}_2, \cdots \bm{\nu}_t$ to obtain a better estimation of the Sliced Cram\'er 2-distance. In this experiment, we set $t=7$, so that $\bm{\nu}_1, \bm{\nu}_2, \cdots \bm{\nu}_7$ form a heptagon.

We also show that our algorithm surpasses the existing gradient descent algorithm, which is descending over the Negative Log Likelihood loss.
\begin{center}
\begin{tabular}{cccc}
    Init & SC2 & NLL & SC2+NLL \\
    \includegraphics[width=0.24\columnwidth]{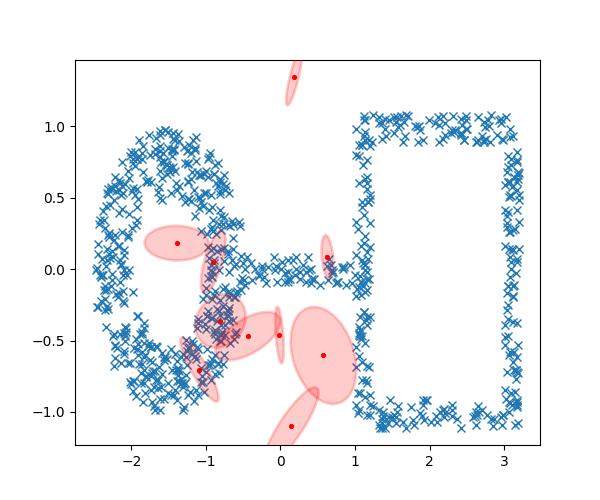} &
    \includegraphics[width=0.24\columnwidth]{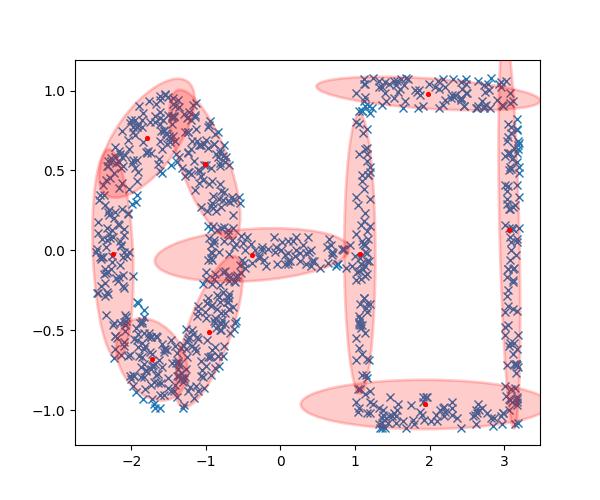} &
    \includegraphics[width=0.24\columnwidth]{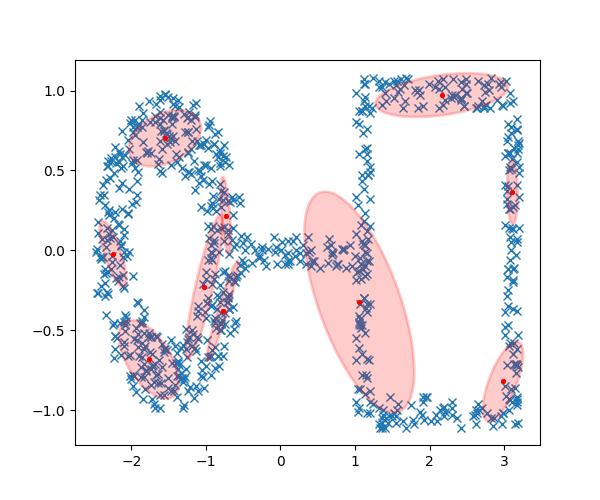} &
    \includegraphics[width=0.24\columnwidth]{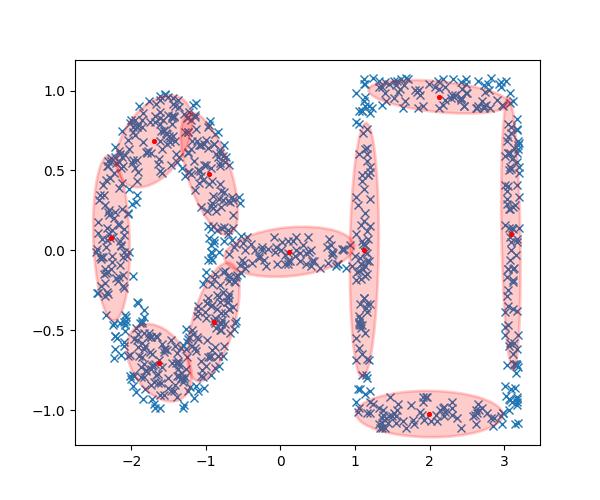} \\
    \includegraphics[width=0.24\columnwidth]{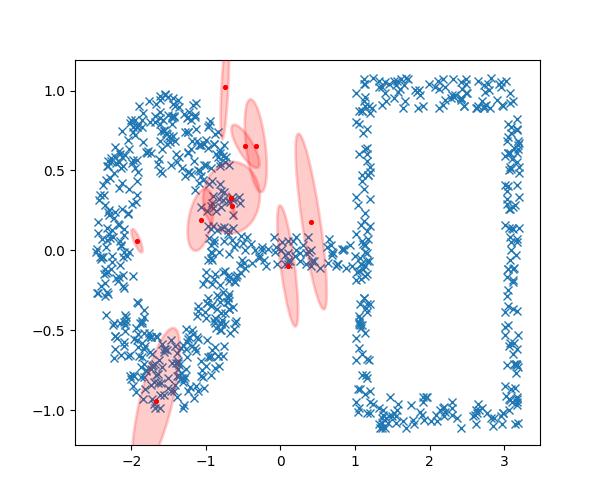} &
    \includegraphics[width=0.24\columnwidth]{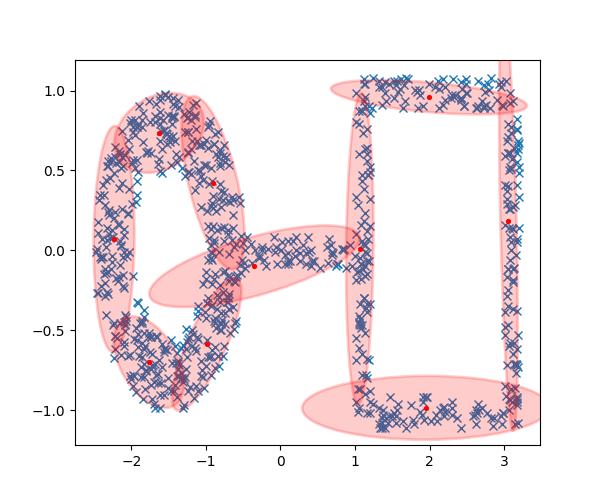} &
    \includegraphics[width=0.24\columnwidth]{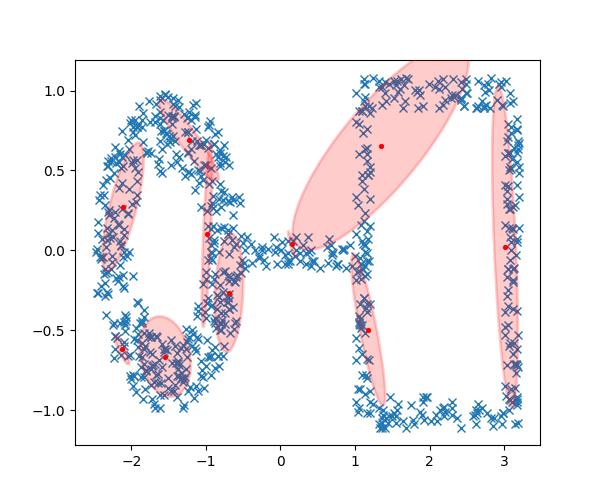} &
    \includegraphics[width=0.24\columnwidth]{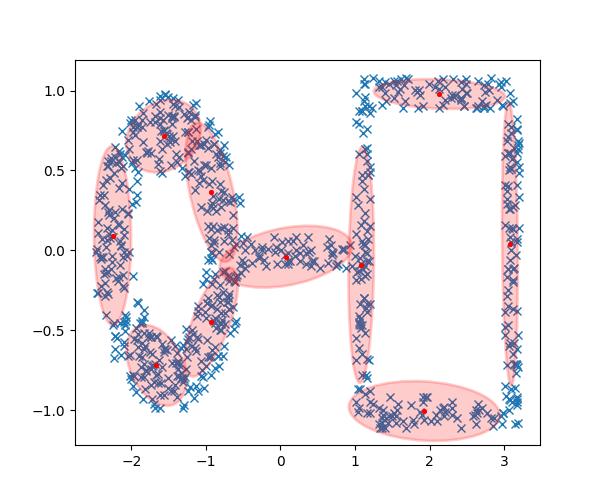} \\
    \includegraphics[width=0.24\columnwidth]{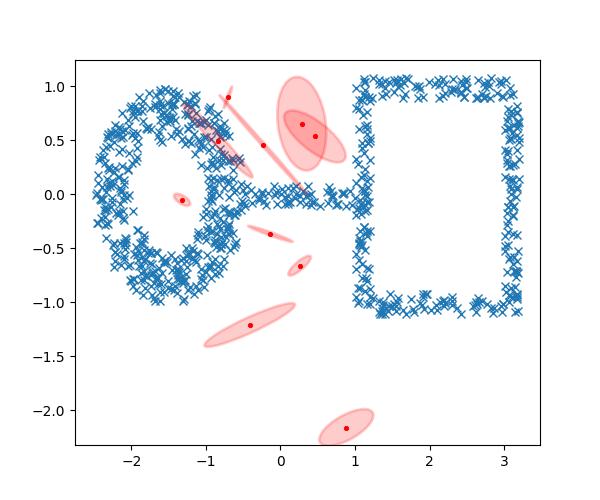} &
    \includegraphics[width=0.24\columnwidth]{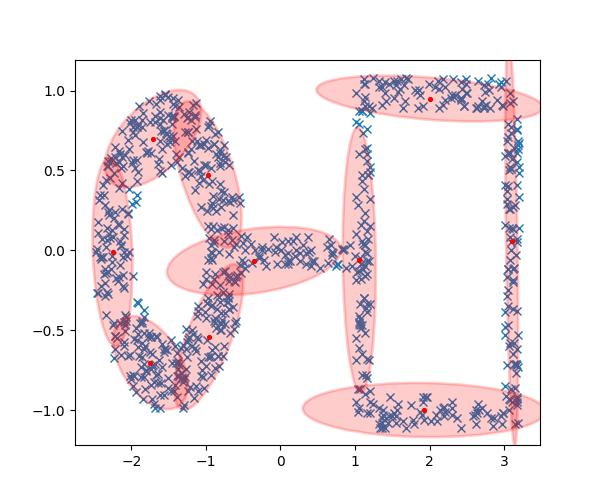} &
    \includegraphics[width=0.24\columnwidth]{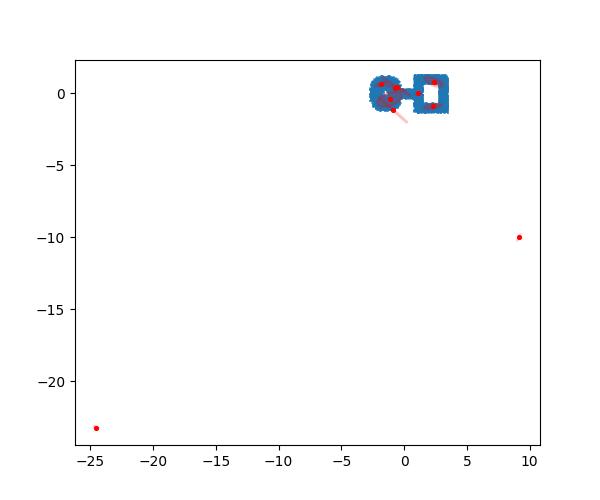} &
    \includegraphics[width=0.24\columnwidth]{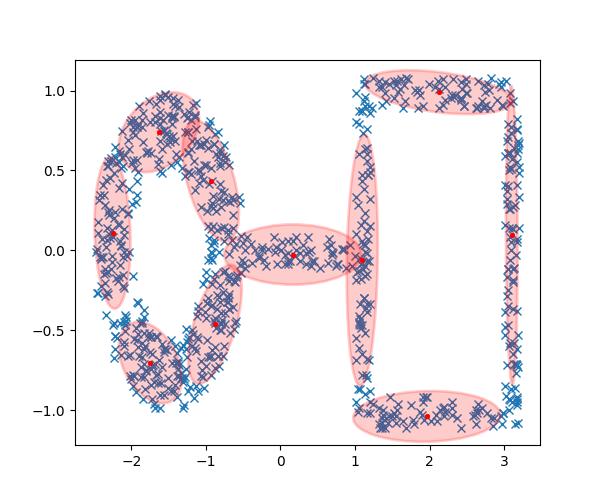}
\end{tabular}
\captionof{figure}{Results. The blue points are the data. Each red ellipse denotes a Gaussian component, whose boundary is the contour of 2 standard deviations.} 
\end{center}
The meaning of each column is explained here:
\begin{itemize}
    \item Init: The initial GMM, without any learning.
    \item SC2: By gradient descent over the Sliced Cram\'er 2-loss for 1200 steps. Learning rates are set to 5e-6, 2e-2, 3e-3 respectively for $\{p_j\}_j,\{\bm{\mu}_j\}_j, \{\bm{\Sigma}_j\}_j$ parts.  
    \item NLL: By gradient descent over the Negative Log Likelihood loss for 1200 steps. The learning rates are the same as the SC2. During this experiment, overflows and underflows are encountered, indicating that this method is numerically unstable.
    \item SC2+NLL: By gradient descent over the Sliced Cram\'er 2-loss for 1200 steps, then gradient descent over the Negative Log Likelihood loss for another 200 steps. The learning rates do not change.
\end{itemize}

As shown in the figure, Pure gradient descent over the Negative Log Likelihood suffers from problems like local minima, degeneration, and instability. Gradient descent over our Sliced Cram\'er 2-loss is generally stable and consistent, yet there are spaces for improvements, since slight overestimations are encountered of the $\{\bm{\Sigma}_j\}$ part.
The best results overall are obtained by "fine-tuning" the results with the NLL loss after the SC2 step, where the overestimations are addressed. 

\begin{center}
    \begin{tabular}{cc}
        \includegraphics[width=0.48\columnwidth]{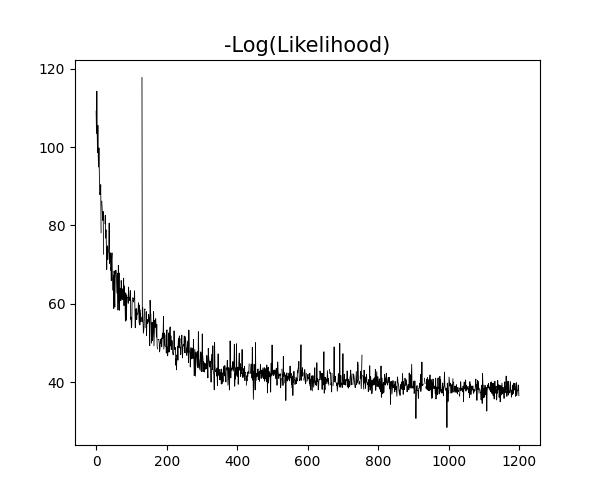} &
        \includegraphics[width=0.48\columnwidth]{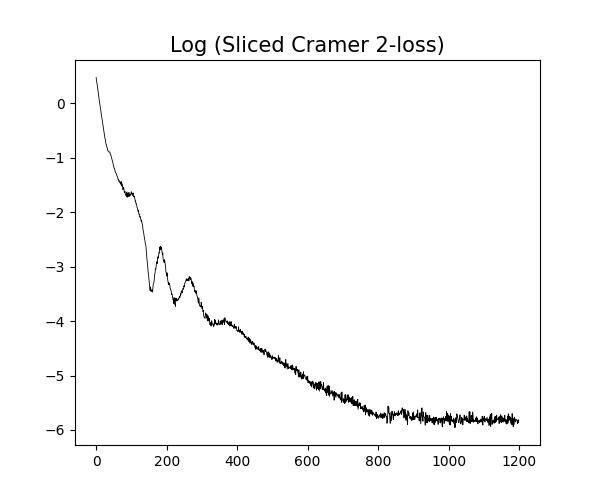}
    \end{tabular}
    \captionof{figure}{Comparison of the two loss functions over steps.} 
    \end{center}

As can be seen from the figure, Sliced Cram\'er 2-loss is much more stable than Negative Log Likelihood loss. Therefore, we recommend only performing the SC2 step, since there is only a slight difference in the results, but the NLL loss is at high risk of instability. It's usually not worth the risk.

\section{Conclusion}\ 

We have successfully proposed the closed formula for Cram\'er 2-loss in the context of univariate GMM learning, as well as the Sliced Cram\'er 2-loss for multivariate GMM learning. Our new methods offer several advantages over previous approaches.

Firstly, our methods, based solely on gradient descent, is particularly beneficial in cases where GMM learning is combined with neural networks. This compatibility allows for easy integration with deep learning libraries and facilitates applications such as training neural networks that output GMMs.

Secondly, our approaches eliminate the need for sampling the target model. By using a loss function between two models, we can directly learn a GMM towards another model, making it possible to apply our methods to tasks like model compression. This expands the range of potential applications and simplifies the learning process.

Additionally, our algorithms come with theoretical guarantees. The loss function is proved to be global Lipschitz for the mean and standard deviation components, preventing gradient explosion, and the sampling gradients are unbiased. These theoretical foundations guarantee that our approach can perform well in various scenarios.

While these are general advantages, there are also more specific advantages to the one-dimensional, univariate case. 

For one thing, the closed-form solution computable by deep learning libraries allows for precise computation of the loss and facilitates the study of its properties. Moreover, our algorithm is directly applicable to Distributional Q-learning, providing both theoretical guarantees and practical convenience. It is parameter-efficient because only a few Gaussian mixtures are required to accurately approximate the continuous and smooth real distribution of $Q$ values commonly encountered in practice. 

Furthermore, our approach enhances interpretability. It completely avoids issues like "zig-zags" (discontinuities) and "crossings" (violations of the monotonicity of the CDF) in the distribution function of QR-DQN and FQF. This enables straightforward computation of Quantiles, Expectiles \cite{rowland2019statistics}, and Conditional-Value-at-Risks (CVaRs) \cite{keramati2020being}.

In summary, our proposed methods provide novel solutions for GMM learning and offer significant advantages, including compatibility with gradient descent, direct learning without sampling, theoretical guarantees, closed-form solutions in the one-dimensional case, applicability in Distributional Q-learning, parameter efficiency, and improved interpretability.

\section{Future work}\ 

In terms of future work, there are several areas that are worthy to explore.

Firstly, conducting more experiments would provide valuable insights. This work primarily focuses on the theoretical foundations and feasibility of our approaches, so only a few simple experiments have been done. It would be beneficial to invite researchers with access to ample computational resources to test our methods on a larger scale, such as the Atari57 benchmark. 

Another area of future research involves investigating numerical stability of the loss function. Although our experiments are not heavily affected by numerical instability issues, it is possible that our algorithms may encounter them, such as \textit{catastrophic cancellations} \cite{cuyt2001remarkable}.
This concern arises from subtracting nearly equal terms in our formula, resulting in a loss of precision. In our experiments in \texttt{float64}, two almost equal terms about $30$ are subtracted, yielding a loss of about $0.003$, which loses approximately $15$ bits of precision. Further study could be conducted to see whether and how this issue would affect performance, and how it could be mitigated.

Additionally, considering the frequent computation of the loss function, it is recommended to optimize the code. One potential optimization strategy is implementing the computation using CUDA or other techniques to make use of parallel processing capabilities and enhance efficiency.

Would you consider integrating this algorithm into your own work, we have the following suggestions:

1. Experiment different learning rates for different parameter sets. It is suggested to set a learning rate for the fractional part, $\{p_j\}$, at most 1/1,000 of the learning rate for $\{\bm{\mu}_j\}$. Differentiation in learning rates helps achieve a balanced optimization process, and avoids degeneration of distribution, since the gradient stability is guaranteed for $\{\bm{\mu}_j\}$ and $\{\bm{\Sigma}_j\}$ components but not for $\{p_j\}$ components.

2. Use higher precision floating point numbers. We suggest at least \texttt{float32} or even \texttt{float64}, to prevent potential problems of catastrophic cancellation. Is also a good practice to use higher precision floating-point types to improve the accuracy and stability of computations.

3. When it's necessary, combine our methods with other techniques, such as the Expectation-Maximization (EM) algorithm, or gradient descent over Negative Log Likelihood loss or Kullback-Leibler divergence to further improve upon results. This combination might help resolve slight overestimations of $\{\bm{\Sigma}_j\}$ component.

By incorporating these suggestions, you might enhance the effectiveness of this algorithm when applying it into your projects.

\newpage
\appendix

\section{Proofs}
\label{appendix:a}
\subsection{Proof of Lemma \ref{lemma:1}}
\begin{proof}
First, the proof that $l_p$ is a metric.
Let $F_1$ and $F_2$ be two functions in $L_{\text{CDF}}^p$. It is easy to show the positivity
$$\int_{-\infty}^{\infty} |F_1(x)-F_2(x)|^p \dd x \ge 0$$
and equality holds iff $F_1 = F_2$ almost everywhere, i.e., except for a zero measure set $S$. 
Let 
$$f(x) = \liminf_{y \in (x, \infty)\backslash S} F_1(y) = \liminf_{y \in (x, \infty)\backslash S} F_2(y)$$
then $f \equiv F_1 \equiv F_2$.

The symmetry is trivial. The triangle inequality is exactly the Minkowski inequality.

\ 

Now we prove that the space $L_{\text{CDF}}^p$ is convex, namely if $F_1$ and $F_2$ are two functions in $L_{\text{CDF}}^p$, then for any $r \in (0,1)$, $rF_1+(1-r)F_2 \in L_{\text{CDF}}^p$.

It is easy to verify by definition that the function $rF_1+(1-r)F_2$ is a CDF. To show that $rF_1+(1-r)F_2 \in L_{\text{CDF}}^p$, we notice by Minkowski's inequality that
$$ \lVert rF_1+(1-r)F_2 - H\rVert _p \le \lVert rF_1 - rH\rVert _p + \lVert  (1-r)F_2 - (1-r)H \rVert _p < \infty $$

This convex property allows us to discuss mixture models.

\ 

Now we prove the completeness: 
Suppose that a sequence of functions $\{F_k\}$ is a Cauchy sequence in $L_{\text{CDF}}^p$, then $\{F_k-H\}$ is a Cauchy sequence in $L^p(\R)$. By the completeness of $L^p$ spaces, we have $F_k-H \to G_0$, where $G_0 \in L^p(\R)$.
Thus, $\lVert F_k - (G_0+H)\rVert _{p} \to 0$. 
We need to find some $F \in L_{\text{CDF}}^p$ such that $F=(G_0+H)$ almost everywhere.

Since $\{F_k\}$ is a sequence converging to $G_0+H$ in $L^p$, there exists a subsequence $\{J_k\}$ such that $\{J_k\}$ converges to $G_0+H$ almost everywhere (details can be found at Theorem 3.9 and 3.12 of the book \cite{rudin1987real}), namely $\R\backslash S$ where $S$ is a zero measure set. 

Let 
$$F(x) = \liminf_{y \in (x, \infty) \backslash S} (G_0+H)(y)$$
Then:
\begin{itemize}
    \item The function $F$ is right continuous and monotonic from the definition.
    \item On $\R\backslash S$, $G_0 + H$ is monotonic: Suppose $a$ and $b$ in $\R\backslash S$ and $a<b$, then $(G_0+H)(a) = \lim_{k \to \infty} J_k(a) \le \lim_{k \to \infty} J_k(b) = (G_0+H)(b)$.
    \item Almost everywhere, $F = G_0 + H$: Since $G_0 + H$ is monotonic on $\R\backslash S$, it is continuous at except countably many points (the set of discontinuous points is denoted by $T$). If $(G_0+H)$ is continuous at $x$, then $F(x) = (G_0+H)(x)$. Therefore, $F = G_0+H$ on $\R \backslash (S\cup T)$, which is almost everywhere.
    \item The limit condition $\lim_{x \to \infty} F(x) = 1$ and $\lim_{x \to -\infty} F(x) = 0$: This is equivalent to proving $\lim_{x \to \pm \infty} (F-H)(x) = 0$. We have $F-H = G_0 \in L^p(\R)$ almost everywhere, and $F-H$ is monotonically increasing and non-positive on $(0,\infty)$. Therefore, $\lim_{x \to infty} F(x) -H(x)$ exists. If $\lim_{x \to infty} F(x) -H(x) = u < 0$, $\int_{0}^{\infty} |F(x)-H(x)|^p \dd x = \infty$, contradiction. Therefore, $\lim_{x \to \infty} F(x) = 1$ and similarly $\lim_{x \to -\infty} F(x) = 0$. 
\end{itemize}
Thus, $F \in L_\text{CDF}^p$ is the limit of $\{F_k\}$, the completeness is proved.
\end{proof}

\subsection{Proof of Lemma \ref{lemma:2}}
\begin{proof}
    We prove that step functions (CDFs of delta mixtures) are dense in the space $L_{\text{CDF}}^p$.

    Suppose $F \in L_{\text{CDF}}^p$, then $F \big|_{(-\infty, 0)} \in L^p(-\infty, 0)$. We construct a series of $\{F_n\} \to F$ with respect to $L^p$ on $(-\infty, 0)$. The part $(0, \infty)$ can be constructed similarly.

    For $n$, let $t_{n,1}, t_{n,2}, \cdots, t_{n, r} = -\frac{2^{2n}}{2^n}, -\frac{2^{2n}-1}{2^n}, \cdots, 0$ respectively. 
    Define
    $$ F_n(x) = \left\{
        \begin{aligned}
            &0, &x<t_{n,1} \\
            &F(t_{n,j}), &x \in [t_{n,j}, t_{n,j+1}) \\
        \end{aligned}
    \right.$$
    Therefore, it's easy to verify that $\{F_n\} \nearrow F$, or $\{F-F_n\} \searrow 0$.
    By monotone convergence theorem, $\{F_n\} \to F \in L^p(-\infty, 0)$. 
    The other part on $(0,\infty)$ can be proved by analogy.
\end{proof}

\subsection{Proof of Theorem \ref{thm:bounded}}
\begin{proof}
    From equation \ref{eqn:c2} we know that
    $$
    \begin{aligned}
        C_2^2(G_1, G_2) &= \sum _{j=1}^n \sum _{k=1}^{n'} \left( p_j p_k' \sqrt{\sigma_j^2+\sigma_k'^2} \cdot  U \left(\frac{\mu_j-\mu_k'}{\sqrt{\sigma_j^2+\sigma_k'^2}}\right) \right)
        + \sum _{j=1}^n \sum _{k=1}^{n'} \left( p_j p_k' \sqrt{\sigma_j^2+\sigma_k'^2} \cdot  U \left(\frac{\mu_k'-\mu_j}{\sqrt{\sigma_j^2+\sigma_k'^2}}\right) \right) \\
        &- \sum _{j=1}^n \sum _{k=1}^n \left( p_j p_k \sqrt{\sigma_j^2+\sigma_k^2} \cdot  U \left(\frac{\mu_j-\mu_k}{\sqrt{\sigma_j^2+\sigma_k^2}}\right) \right) 
        - \sum _{j=1}^{n'} \sum _{k=1}^{n'} \left( p_j' p_k' \sqrt{\sigma_j'^2+\sigma_k'^2} \cdot  U \left(\frac{\mu_j'-\mu_k'}{\sqrt{\sigma_j'^2+\sigma_k'^2}}\right) \right) \\
    \end{aligned}
    $$
    WLOG, let $j=1$. Take partial derivative of $\mu_1$:
    $$
    \begin{aligned}
        \frac {\partial (C_2^2(G_1, G_2))} {\partial \mu_1} 
        &= \sum _{k=1}^{n'} \left( p_1 p_k' \cdot  \frac{\dd U}{\dd x} \left(\frac{\mu_1-\mu_k'}{\sqrt{\sigma_1^2+\sigma_k'^2}}\right) \right)
        + \sum _{k=1}^{n'} \left( -p_1 p_k' \cdot  \frac{\dd U}{\dd x} \left(\frac{\mu_k'-\mu_1}{\sqrt{\sigma_1^2+\sigma_k'^2}}\right) \right) \\
        &- \sum _{k=2}^n \left( p_1 p_k \cdot \frac{\dd U}{\dd x} \left(\frac{\mu_1-\mu_k}{\sqrt{\sigma_1^2+\sigma_k^2}}\right) \right) 
        - \sum _{j=2}^n \left( -p_j p_1 \cdot \frac{\dd U}{\dd x} \left(\frac{\mu_j-\mu_1}{\sqrt{\sigma_j^2+\sigma_1^2}}\right) \right)
    \end{aligned}
    $$
    Since $|{\dd U}/{\dd x}| = |\Phi(x)| < 1$, we have
    $$\left|\frac {\partial (C_2^2(G_1, G_2))} {\partial \mu_1}\right| \le \sum _{k=1}^{n'} p_1 p_k' + \sum _{k=1}^{n'} p_1 p_k' + \sum _{k=2}^n p_1 p_k + \sum _{j=2}^n p_j p_1 \le 4$$

    To show that $$\left|\frac {\partial (C_2^2(G_1, G_2))} {\partial \sigma_1}\right| \le 4 $$ it suffices to show that 
    $$\left|\frac {\partial} {\partial \sigma_1} \left( \sqrt{\sigma_1^2+s^2} \cdot  U \left(\frac{h}{\sqrt{\sigma_1^2+s^2}} \right) \right) \right| \le 1, \quad \forall h, s \in \R $$
    Let $z=\sqrt{\sigma_1^2+s^2}$, then 
    $$\left|\frac {\partial z} {\partial \sigma_1}\right| = \left|\frac{\sigma_1}{\sqrt{\sigma_1^2+s^2}}\right| \le 1$$
    We have
    $$\left|\frac {\partial} {\partial z}\left( z \cdot  U \left(\frac{h}{z} \right)\right)\right| = \left| U \left(\frac{h}{z} \right) - \frac{h}{z} \Phi \left(\frac{h}{z} \right) \right| = \frac{1}{\sqrt{2\pi}} \exp\left(-\frac{h^2}{2z^2}\right) < 1$$
    So 
    $$\left|\frac {\partial} {\partial \sigma_1} \left( \sqrt{\sigma_1^2+s^2} \cdot  U \left(\frac{h}{\sqrt{\sigma_1^2+s^2}} \right) \right) \right| \le \left|\frac {\partial z} {\partial \sigma_1}\right| \cdot \left|\frac {\partial} {\partial z}\left( z \cdot  U \left(\frac{h}{z} \right)\right)\right| \le 1$$
    Therefore we have proved
    $$ \left| \frac {\partial L} {\partial \mu_1} \right| \le 4, \quad \left| \frac {\partial L} {\partial \sigma_1} \right| \le 4.$$ 

\end{proof}

\subsection{Proof of Theorem \ref{thm:sc2prop}}
\begin{proof}
    Proof mainly from \cite{bellemare2017cramer}. We use the equivalence between the Cram\'er 2-distance and the Energy distance \cite{rizzo2016energy} in the univariate case, which means that for any independent random variables $Z, Z' \sim P$ and $W, W' \sim Q$, 
    $$2C_2^2(P, Q) = \mathcal{E}(P,Q) =  2\E[|Z-W|] - \E[|Z-Z'|] - \E[|W-W'|]$$
    \begin{itemize}
        \item Independent sum: Let $\bm{X}'\sim \bm{X}$, $\bm{Y}'\sim \bm{Y}$, $\bm{A}'\sim \bm{A}$ be independent copies of $\bm{X}$, $\bm{Y}$, $\bm{A}$ respectively. Then 
        $$ 
        \begin{aligned}
            & 2S_2^2(\bm{A}+\bm{X}, \bm{A}+\bm{Y}) \\
            &= \int_{\bm{\nu} \in \mathbb{S}^{m-1}} \left(2\E[|\langle \bm{A}+\bm{X}-\bm{A}-\bm{Y}, \bm{\nu} \rangle|] - \E[|\langle \bm{A}+\bm{X}-\bm{A}'-\bm{X}', \bm{\nu} \rangle|] - \E[|\langle \bm{A}+\bm{Y}-\bm{A}'-\bm{Y}', \bm{\nu} \rangle|]\right) \dd \bm{\nu} \\
            &\le \int_{\bm{\nu} \in \mathbb{S}^{m-1}} \left(2\E[|\langle \bm{X}-\bm{Y}, \bm{\nu} \rangle|] - \E[|\langle \bm{X}-\bm{X}', \bm{\nu} \rangle|] - \E[|\langle \bm{Y}-\bm{Y}', \bm{\nu} \rangle|]\right) \dd \bm{\nu} \\
            &= 2S_2^2(\bm{X}, \bm{Y}) \\
        \end{aligned}
        $$
        Where the inequality is primarily due to $|a+b| \ge |b| + a\cdot \sgn (b)$, and $\bm{A}-\bm{A}'$, $\bm{X}-\bm{X}'$, $\bm{Y}-\bm{Y}'$ are independent. 

        \item Scaling property:
        $$ 
        \begin{aligned}
            2S_2^2(c\bm{X}, c\bm{Y}) &= \int_{\bm{\nu} \in \mathbb{S}^{m-1}} \left(2\E[|\langle c\bm{X}-c\bm{Y}, \bm{\nu} \rangle|] - \E[|\langle c\bm{X}-c\bm{X}', \bm{\nu} \rangle|] - \E[|\langle c\bm{Y}-c\bm{Y}', \bm{\nu} \rangle|]\right) \dd \bm{\nu} \\
            &= c\int_{\bm{\nu} \in \mathbb{S}^{m-1}} \left(2\E[|\langle \bm{X}-\bm{Y}, \bm{\nu} \rangle|] - \E[|\langle \bm{X}-\bm{X}', \bm{\nu} \rangle|] - \E[|\langle \bm{Y}-\bm{Y}', \bm{\nu} \rangle|]\right) \dd \bm{\nu} \\
            &= 2cS_2^2(\bm{X}, \bm{Y}) \\
        \end{aligned}
        $$

        \item Unbiased sampling gradient: Suppose $\bm{Y} \sim G_\theta$ and $\hat{\bm{X}} \sim \hat{P}$. Let $\hat{\bm{X}}'$ and $\bm{Y}'$ be independent copies of $\hat{\bm{X}}$ and $\bm{Y}$ respectively.
        $$ 2S_2^2(\hat{\bm{X}}, \bm{Y}) = \int_{\bm{\nu} \in \mathbb{S}^{m-1}} \left(2\E[|\langle \hat{\bm{X}}-\bm{Y}, \bm{\nu} \rangle|] - \E[|\langle \hat{\bm{X}}-\hat{\bm{X}}', \bm{\nu} \rangle|] - \E[|\langle \bm{Y}-\bm{Y}', \bm{\nu} \rangle|]\right) \dd \bm{\nu} $$
        The gradient of the sample loss with respect to parameter $\theta$:
        $$ \nabla_\theta \left(2S_2^2(\hat{\bm{X}}, \bm{Y})\right) = \int_{\bm{\nu} \in \mathbb{S}^{m-1}} \left(2\nabla_\theta \E[|\langle \hat{\bm{X}}-\bm{Y}, \bm{\nu} \rangle|] - \nabla_\theta \E[|\langle \bm{Y}-\bm{Y}', \bm{\nu} \rangle|]\right) \dd \bm{\nu} $$
        The gradient of the true loss with respect to parameter $\theta$:
        $$ \nabla_\theta \left(2S_2^2(\bm{X}, \bm{Y})\right) = \int_{\bm{\nu} \in \mathbb{S}^{m-1}} \left(2\nabla_\theta \E[|\langle \bm{X}-\bm{Y}, \bm{\nu} \rangle|] - \nabla_\theta \E[|\langle \bm{Y}-\bm{Y}', \bm{\nu} \rangle|]\right) \dd \bm{\nu} $$
        It suffices to show that
        $$ \int_{\bm{\nu} \in \mathbb{S}^{m-1}} \nabla_\theta \E[|\langle \bm{X}-\bm{Y}, \bm{\nu} \rangle|] \dd \bm{\nu} = \E_{\mathcal{X}}\left[\int_{\bm{\nu} \in \mathbb{S}^{m-1}} \nabla_\theta \E_{\hat{\bm{X}} \sim \hat{P}} [|\langle \hat{\bm{X}}-\bm{Y}, \bm{\nu} \rangle|] \dd \bm{\nu}\right]$$
        By commutativity of integrals, we have
        $$\E_{\mathcal{X}}\left[\int_{\bm{\nu} \in \mathbb{S}^{m-1}} \nabla_\theta \E_{\hat{\bm{X}} \sim \hat{P}} [|\langle \hat{\bm{X}}-\bm{Y}, \bm{\nu} \rangle|] \dd \bm{\nu}\right] = \int_{\bm{\nu} \in \mathbb{S}^{m-1}} \nabla_\theta \E_{\mathcal{X}}\left[ \E_{\hat{\bm{X}} \sim \hat{P}} [|\langle \hat{\bm{X}}-\bm{Y}, \bm{\nu} \rangle|] \right] \dd \bm{\nu}$$
        By simplification 
        $$\E_{\mathcal{X}}\left[ \E_{\hat{\bm{X}} \sim \hat{P}} [|\langle \hat{\bm{X}}-\bm{Y}, \bm{\nu} \rangle|] \right] = \E_{\bm{a} \sim P} [|\langle \bm{a}-\bm{Y}, \bm{\nu} \rangle|] = \E_{\bm{X} \sim P} [|\langle \bm{X}-\bm{Y}, \bm{\nu} \rangle|]$$

        Moreover, if we rewrite the integral operator $\int_{\bm{\nu} \in \mathbb{S}^{m-1}}$ as the expectation operator $B_{m-1} \E_{\bm{\nu} \in \mathbb{S}^{m-1}}$, then 
        $$\begin{aligned}
            \nabla_\theta S_2^2 (G_\theta, P) &=\E _{\mathcal{X}\sim P} \left[\nabla_\theta S_2^2 (G_\theta, \hat{P})\right] \\
            &= \int_{\bm{\nu} \in \mathbb{S}^{m-1}} \E _{\mathcal{X}\sim P} \left[\nabla_\theta C_2^2 (\langle G_\theta, \bm{\nu}\rangle, \langle \hat{P}, \bm{\nu}\rangle)\right] \dd \bm{\nu}\\
            &= B_{m-1}\cdot \E _{\bm{\nu}} \E _{\mathcal{X}\sim P} \left[\nabla_\theta C_2^2 (\langle G_\theta, \bm{\nu}\rangle, \langle \hat{P}, \bm{\nu}\rangle)\right] 
        \end{aligned}$$
        and we have proved all three properties.
    \end{itemize}
\end{proof}

\subsection{Proof of Theorem \ref{thm:boundedsc2}}
\begin{proof}

    For the first part, replace $\bm{\mu}_1$ by $\bm{\mu}_1+\alpha \bm{\lambda}$ where $\lVert\bm{\lambda}\rVert\le 1$. 
    We prove that $$\left \lvert\frac{\partial L}{\partial \alpha}\right\rvert _{\alpha=0} \le 4B_{m-1}$$
    We only need to show that 
    $$\left\lvert\frac{\partial C_2^2(\langle G_1, \bm{\nu} \rangle, \langle G_2, \bm{\nu} \rangle)}{\partial \alpha}\right\rvert_{\alpha=0} \le 4$$
    Since $$\left\lvert\frac{\partial C_2^2(\langle G_1, \bm{\nu} \rangle, \langle G_2, \bm{\nu} \rangle)}{\partial \alpha}\right\rvert = \left\lvert\frac{\partial C_2^2(\langle G_1, \bm{\nu} \rangle, \langle G_2, \bm{\nu} \rangle)}{\partial \langle\bm{\mu}_1+\alpha \bm{\lambda}, \bm{\nu}\rangle}\right\rvert \cdot \left\lvert\frac{\partial \langle\bm{\mu}_1+\alpha \bm{\lambda}, \bm{\nu}\rangle}{\partial \alpha}\right\rvert$$
    By Theorem \ref{thm:bounded}, $$\left\lvert\frac{\partial C_2^2(\langle G_1, \bm{\nu} \rangle, \langle G_2, \bm{\nu} \rangle)}{\partial \langle\bm{\mu}_1+\alpha \bm{\lambda}, \bm{\nu}\rangle}\right\rvert_{\alpha=0} \le 4$$
    and obviously $$\left\lvert\frac{\partial \langle\bm{\mu}_1+\alpha \bm{\lambda}, \bm{\nu}\rangle}{\partial \alpha}\right\rvert \le 1$$

    \ 

    For the second part, replace $\bm{S}_1$ by $\bm{S}_1+\beta \bm{R}$ where $\lVert\bm{R}\rVert\le 1$. Here the norm is the $\mathit{l}_2$ norm of matrices, namely $\lVert\bm{A}\rVert = \sqrt{\tr(\bm{A}\T\bm{A})}$.
    
    We prove that $$\left \lvert\frac{\partial L}{\partial \beta}\right\rvert _{\beta=0} \le 4B_{m-1}$$
    We only need to show that 
    $$\left\lvert\frac{\partial C_2^2(\langle G_1, \bm{\nu} \rangle, \langle G_2, \bm{\nu} \rangle)}{\partial \beta}\right\rvert_{\beta=0} \le 4$$
    Since $$\left\lvert\frac{\partial C_2^2(\langle G_1, \bm{\nu} \rangle, \langle G_2, \bm{\nu} \rangle)}{\partial \beta}\right\rvert = 
    \left\lvert\frac{\partial C_2^2(\langle G_1, \bm{\nu} \rangle, \langle G_2, \bm{\nu} \rangle)}{\partial \lVert(\bm{S}_1+\beta\bm{R})\bm{\nu}\rVert}\right\rvert 
    \cdot \left\lvert\frac{\partial \lVert(\bm{S}_1+\beta\bm{R})\bm{\nu}\rVert}{\partial \beta}\right\rvert$$
    Where we have $\sigma_1 = \lVert(\bm{S}_1+\beta\bm{R})\bm{\nu}\rVert$.
    By Theorem \ref{thm:bounded}, $$\left\lvert\frac{\partial C_2^2(\langle G_1, \bm{\nu} \rangle, \langle G_2, \bm{\nu} \rangle)}{\partial \lVert(\bm{S}_1+\beta\bm{R})\bm{\nu}\rVert}\right\rvert _{\beta=0} \le 4$$
    and $\lVert \bm{R} \rVert \le 1$, $\lVert \bm{\nu} \rVert \le 1$, so $\lVert \bm{R}\bm{\nu} \rVert \le 1$.
    Which yields $$\left\lvert\frac{\partial \lVert\bm{S}_1\bm{\nu}+\beta\bm{R}\bm{\nu}\rVert}{\partial \beta}\right\rvert \le 1$$
\end{proof}

\section{Implementation of the Cram\'er 2-distance Function}
\label{appendix:b}
Below is the implementation of the Cram\'er 2-distance function in Python.
\begin{minted}{python} 
import torch
import torch.nn as nn
import torch.nn.functional as F
class CramerUnit(nn.Module):
    def __init__(self):
        super().__init__()
        # 0.797884560802865356 = sqrt(2/pi)
        self.unit = lambda z: 2 * F.gelu(z) - z + 0.797884560802865356 * torch.exp(-z**2/2)
    def forward(self, m1, s1, m2, s2):
        v = torch.sqrt(s1**2 + s2**2 + 1e-20) 
        return v * self.unit((m1-m2) / v)
\end{minted}
\end{document}